\documentclass[11pt]{article}

\usepackage{EACL2023}

\usepackage{times}
\usepackage{latexsym}

\usepackage[T1]{fontenc}

\usepackage[utf8]{inputenc}

\usepackage{microtype}

\usepackage{inconsolata}

\usepackage{amsthm}
\usepackage{amsmath}
\usepackage{amsfonts}
\usepackage{amssymb}
 \usepackage{mathtools} 
\usepackage{caption}
\usepackage{subcaption}
\usepackage{dsfont}
\usepackage{mathtools}
\usepackage{bbm}
\usepackage{bm}
\usepackage{multirow}
\usepackage{multicol}
\usepackage{natbib}
\usepackage{tikz}
\usepackage{amssymb,amsmath,linguex,amsthm}
\theoremstyle{definition}
\newtheorem{definition}{Definition}

\newtheorem{theorem}{Theorem}
\newtheorem{prop}{Proposition}
\newtheorem{cor}{Corollary}

\newcommand{\open}{\mathcal{O}}

\newcommand{\hidden}[1]{}

\let\emptyset\varnothing

\begin{document}
\title{Limits for learning with Language Models}

\author{Nicholas Asher \\ CNRS, IRIT \\\texttt{asher@irit.fr}  
    \And 
  Swarnadeep Bhar \\ Université de Toulouse 3 \\IRIT  \\\texttt{swarnadeep.bhar@irit.fr}  
  \And 
  Akshay Chaturvedi\\ Université de Toulouse 3 \\ IRIT \\\texttt{akshay91.isi@gmail.com}  
  \\
  \AND 
Julie Hunter \\LINAGORA \\\texttt{jhunter@linagora.com}
  \And
  Soumya Paul \\ TELINDUS\\ \texttt{soumya.paul@gmail.com}}




\maketitle
\begin{abstract}
With the advent of large language models (LLMs), the trend in NLP has been to train LLMs on vast amounts of data to solve diverse language understanding and generation tasks.  The list of LLM successes is long and varied. Nevertheless, several recent papers provide empirical evidence that LLMs fail to capture important aspects of linguistic meaning. Focusing on universal quantification, we provide a theoretical foundation for these empirical findings by proving that LLMs cannot learn certain fundamental semantic properties including semantic entailment and consistency as they are defined in formal semantics.  More generally, we show that LLMs are unable to learn concepts beyond the first level of the Borel Hierarchy, which imposes severe limits on the ability of LMs, both large and small, to capture many aspects of linguistic meaning.  This means that LLMs will continue to operate without formal guarantees on tasks that require entailments and deep linguistic understanding.

\end{abstract}

\maketitle
\section{Introduction}

The success of large language models (LLMs) has led researchers in NLP to harness LLMs trained on vast amounts of data to solve a variety of language understanding and generation tasks, and some have claimed that LLMs can solve any task that can 
be specified via prompting \cite{brown:etal:2020}.  While the list of LLM successes is long, there have been several recent papers that provide empirical evidence that LLMs at least sometimes fail to capture important aspects of linguistic meaning \cite{kuhnle:copestake:2019,sinha:etal:2020unnatural,yuksekgonul:etal:2022,chaturvedi:etal:2022,kalouli:etal:2022}.  Those who have dabbled in ``BERTology'' with respect to linguistic meaning often have the feeling that fixing one LLM deficiency just leads to the discovery of new ones.  

This paper provides a theoretical explanation of certain of these observed failings of LLMs.  In particular, we prove that LLMs cannot learn the notions of semantic entailment or consistency as defined in formal semantics \cite{dowty:etal:1981} because they are incapable of mastering universal quantification. Our work builds on \citet{siegelmann:sontag:1992,siegelmann:2012,weiss:etal:2018}, concerning the expressive power of neural networks, 
but we focus on the learnability of semantic concepts and use novel tools.

Our argument has widespread implications: not only does a general capacity to recognize semantic entailment and consistency underlie everyday conversational interactions, but the meanings of a great many common linguistic expressions  depend on universal quantification. This set includes---but is certainly not limited to---a long list of quantifiers ({\em every, some, many, most,...  every other}, ...),  temporal adverbs ({\em always, never, eventually}) that are essential to planning \cite{lamport:1980}, modal operators (\textit{possibly}, \textit{necessarily},...), and certain discourse connectives and adverbs (\textit{therefore}, \textit{if / then}, \textit{except}, \textit{because}, ...).  

We begin in Section \ref{sec:context} by contextualizing our claims in terms of expectations about the linguistic capacities and applications of LLMs.  In Section \ref{sec:continuations}, we introduce the framework of continuation semantics, which will allow us to adapt certain notions central to truth-conditional semantics to the case of LLMs. Section \ref{sec:limits} lays out the core of our theoretical argument, focusing first on what is needed to learn universal quantification and then  generalizing our argument to a wide range of linguistic expressions. Our theoretical argument suggests that we should expect certain empirical failures from LLMs, and in Section \ref{sec:q1}, we provide  evidence that our predictions are borne out.  Section \ref{sec:conclusion} concludes. 

\section{Context}\label{sec:context}

Our results are particularly relevant to downstream tasks that  require an agent to not only create fluent, creative and contextually relevant speech but also to act precisely based on the meaning of linguistic expressions and reliably recognize semantic inconsistency. For a robot that has been instructed (via conversation) to tighten {\em every} screw of a door, to {\em never} walk on an airplane wing, or to \textit{stop} drilling \textit{immediately if} certain conditions hold, acting appropriately requires being able to infer what do to based on the \textit{linguistic meaning} of the words {\em every}, \textit{never}, \textit{stop}, \textit{immediately} and \textit{if}---and in these cases, getting things  \textit{mostly} right won't do, especially if lives or substantial economic loss are at risk. 

An important corollary of our argument is that while it might be tempting to separate reasoning and linguistic competence \citep{mahowald:etal:arxiv}, the former is in fact inextricably tied to our ability to draw inferences based on \textit{linguistic} content---not just on, say, mathematical or real-world facts.   This in turn suggests that approaches which attempt to patch up  knowledge deficiencies for LLMs by giving them  access to external models \citep{mialon2023augmented} will fall short in developing reliable models of linguistic understanding because LLMs fail to grasp the notions that underlie the very way that sentences (and actions) are woven together in conversation. 

Empirical studies like \citet{chaturvedi:etal:2022} show that LLM failures to respect semantic entailment in question answering tasks follow from fundamental features of LLM training; thus while extensive training and large data sets may improve LLM results,  performance will inevitably remain unstable and we should continue to expect hallucinations and reasoning errors in NLP tasks like question-answering and natural language inference.

 \section{Language models and formal semantics with continuations}\label{sec:continuations}

\subsection{LLMs and strings}
 We consider LLMs trained on transformer architectures over very large corpora using classic language modeling tasks, namely masked language modeling or next sentence prediction.  The former involves masking certain words in a given corpus and training the model to guess the missing words, while in the latter, a context (a sentence typically) is provided to the model, which is trained to predict the sentence that follows. 
This unsupervised training allows language models to build rich internal representations that have been shown through probing to contain at least implicitly a large amount of linguistic information \cite{bert,roberta,tenney:etal:2018}. 

Formally, LLMs learn a function $f: C \times X \rightarrow [0,1]$ that assigns a probability to a word (or string or discourse move)  $x \in X$ given a context (or finite string) $C$.  More abstractly, let $V$ be a countable set called the vocabulary.  For $i>0$, let $V^i$ denote the set of all length $i$ strings in the vocabulary $V$ and $V^{\leq i}$ denote the set of all strings $V$ whose length is at most $i$. $V^*$ denotes the set of all finite strings and $V^\omega$ the set of countably infinite strings in $V$. We can then rewrite $f$ as $f \colon V^{\leq n}  \rightarrow \mu$, where $\mu$ is a probability measure (which is often called its \emph{prediction}) over $V^{n+m}$ for $m \geq 1$. Typically, the prediction function is used on strings of length $m$ where $m$ is smaller than $n$.  
 
 By exploiting $f$, an LLM can extend $\mu$ to a distribution on the set of strings $V^*$.   The most straightforward way is to follow autoregressive models that calculate the probability of strings via conditionalization. For a new sentence $s' = (w_1, w_2, ... , w_{m+1})$, and an input string $s$ of length n provided as context, we have:  
\begin{equation}\label{calculation} \mu^{n+m+1}(s' | s ) = \mu^{n+1}(w_1 | s)\times \mu^{n+2}(w_2 \ | s,w_1) \times 
\end{equation}
$$ ... \times \mu^{n+m}(w_n \ | s,w_{m-1},...,w_1)$$
For any $s'\in V^*,\ \mu(s')$ represents the confidence with which an LLM predicts $s'$, after training on strings in $V^{\leq n}$.

\subsection{Linguistic meaning}
In what follows, we are in interested strings that have a well formed meaning and are evaluable as true or false.  Linguists use truth conditional semantics to define the meanings of strings or well formed sentences in terms of the conditions under which they are true.  Thanks to the work of \citet{tarski:1944,tarski:1936}, we can formalize the notion of truth conditions using the set-theoretic notion of a model that defines denotations or truth conditions for sentences recursively from denotations for sentential constituents \cite{dowty:etal:1981}. 

The notion of a model not only serves to define truth conditions; it also captures entailments.   
We define the notion of \textit{semantic consequence} using the notion of a model or structure $\mathfrak{A}$ as follows \citep{chang:keisler:1973}: 
\begin{definition}
$\phi$ is a semantic consequence of $\Gamma$ (in symbols, $\Gamma \models \phi$) if and only if in every structure $\mathfrak{A}$ in which $\Gamma$ is satisfied ($\mathfrak{A} \models \Gamma$), $\mathfrak{A}$ also makes  true or satisfies $\phi$ ($ \mathfrak{A}\models \phi$). That is: $ \forall  \mathfrak{A}, \ \mathfrak{A} \models \Gamma \ \Rightarrow \  \mathfrak{A}\models \phi$
\end{definition}
The notion of semantic consequence integrates entailment with truth conditional meaning; two strings have exactly the same entailments just in case they are true in the same models.  Accordingly we can capture the truth conditional meaning of a string in terms of the strings that it entails.  \textit{Socrates is a man}, for example, entails \textit{Socrates is human}, \textit{Socrates is mortal}, \textit{Socrates is an adult} but also that \textit{someone} is a man, human, mortal and so on. What it means for Socrates to be a man (and, indirectly, the meaning of \textit{man}) can be captured by the full set of these entailments.

Our idea is to apply truth conditional semantics to LLMs by representing models themselves as strings.  Semanticists have used strings and {\em continuation semantics} \citep{reynolds:1974} ---in which the meaning of a string $s$ is defined in terms of its possible \textit{continuations}, the set of longer strings $S$ that contain $s$---to investigate the meaning and strategic consequences of conversational moves \cite{JPL}, temporal expressions \cite{fernando:2004}, generalized quantifiers \cite{graf:2019}, and the ``dynamic'' formal semantics of \cite{kamp:reyle:1993,asher:1993}\citep{degroote:2006,asher:pogodalla:2011}. In our case, we will use strings to define models ${\mathfrak A_s}$.  We will use this trick to reformulate semantic consequence: where  $\|\phi\|$ is the set of strings describing models that satisfy a truth evaluable string $\phi$:$\Gamma \models \phi$ iff $\|\Gamma \| \subseteq \|\phi\|$.




LLMs naturally find their place in such a framework \cite{fernando:2022}: given their training regime, the meaning of any natural language expression for an LLM is  a function from input contexts to sets of larger strings or continuations.  LLMs provide a probability distribution over possible continuations and can predict possible continuations of a given text or discourse.

\section{Learning limits for semantic concepts}\label{sec:limits}

Semantic consequence defines linguistic entailments and importantly provides the fundamental connection between meaning and inference that ensures linguistic understanding \cite{montague:1974}. Crucial to $\models$ is the use of universal quantification over all possible structures---an infinite space of possible circumstances of evaluation or set of possibilities. A true grasp of semantic consequence thus requires an understanding of universal quantification at least over countably infinite domains.  In Section \ref{subsec:every}, we show that an LLM's training regime makes it fundamentally unable to learn the concept of universal quantification. In Section \ref{subsec:general}, we generalize our argument to show that LLMs are incapable of learning a wide variety of everyday semantic concepts.

\subsection{Learning the full meaning of {\em every}}\label{subsec:every}


To see if the set of strings that define the concept \textit{every} is learnable for an LLM, consider \ref{ex:every}.

\ex.\label{ex:every} Every object is blue.

We will use strings of atomic formulas and their negations to define models (or more precisely their atomic diagrams) that we will use to test whether an LLM $M$ can learn the concept of universal quantification through inductive reasoning from a series of individual trials over finite subsequences 
of strings representing countably infinite domains.   
 In particular, we will ask whether it is possible to train an LLM $M$ to judge, for a string $s$ of arbitrary length $n$, whether $s$ is consistent with \ref{ex:every}, or equivalently, given that $s$ defines a model $\mathfrak{A}_s$, whether given $\mathfrak{A}_s$, \ref{ex:every} is true.  If $M$ can reliably judge in which models $\mathfrak{A}_s$ \ref{ex:every} is true, we can conclude it has learned the meaning of {\em every}.

To this end, consider a language ${\cal L}$ containing negation, the predicate {\em is blue} and a countably infinite number of constants $a_i$ enumerating objects of a countably infinite domain.   ${\cal L} $ formulas are of the  form $a_i$ {\em is blue} and $a_i$ {\em is not blue}.  
 We use the formulas of ${\cal L}$ as ``words'' to define the set of finite strings, $V_{\cal L}^*$ and the set of countably infinite strings $V_{\cal L}^\omega$.  Each such string corresponds to a finite or countably infinite model in which \ref{ex:every} is true or not.  
 

 In the course of training, $M$ will be presented with finite sequences that define structures of increasing size.  For each $n$ and set of models of size $n$, $M$ will form a set of hypotheses ${\cal H}^n$, where for $h \in {\cal H}^n, h: V^n_{\cal L} \rightarrow \{0,1\}$.  ${\cal H}^n$ corresponds to the hypothesis space of the problem; each $h^n$ says whether a presented sequence of length $n$ is consistent with \ref{ex:every}.  As each $h^n \in {\cal H}^n$ is a characteristic function of a subset of $V^n_{\cal L}$, we can identify hypotheses with sets of strings.  So for instance, $h^n_\forall$ is the set of strings in $V^n_{\cal L}$ that are consistent with \ref{ex:every} and that define models in which \ref{ex:every} is true.  We will additionally assume that  $h^n_\forall$  picks out a suitable set for each $V^n$, and similarly for each $h^n_k$.  

However simply learning $h^n_\forall$ for some $n$ will not be sufficient for $M$ to learn the meaning of {\em every}.  Universal quantification is a concept that applies to arbitrarily large domains.  So the question, {\em Can $M$ inductively learn the meaning of {\em every}?} becomes {\em Can $M$ inductively learn hypothesis $h^\omega_\forall \in {\cal H}^\omega$?} 

To answer this question, we first have to specify what we mean by inductive learning.  Recall that an LLM $M$ has learned from unsupervised training a function $f \colon V^{\leq n} \rightarrow \mu(V^{\leq n+m})$ with $\mu(V^{\leq n+m})$ a probability distribution over completions of length $m$ of  contexts of length $n$. An LLM can use this distribution to compute probability values for arbitrarily long strings or continuations using Equation \ref{calculation}.  

In the case at hand, $M$ needs to use this distribution over ${\cal L}$ strings to compute the probability that a string $s$ is in $h_\forall$ or the probability of $s$ given $h_\forall$.   
To learn inductively $M$ must use its training data $D^{\leq n}$ to update its prior for the distribution $\mu^{n}$ using a rationally justifiable form of {\em inductive inference}; e.g., for $h \in {\cal H}^{n}, \mu^{n}(s|h) = \frac{\mu^{n}(h | s) \times \mu^{n}(s)}{\mu^{ n}(h)}$.   

Additionally, we consider two constraints on distributions to define learning in terms of an inductively inferred change in the distribution from the priors.  The first constraint, {\em Max Ent}, says that the distribution $\mu$ prior to training should assign all hypotheses a weight based on maximum entropy or a least informative distribution. This is usual with auto-regressive models and a common assumption in other models.  

The second constraint is that distributions for inductive learning should be {\em non-degenerate}. We have assumed that our LLM $M$ has been trained over sequences of length $n$. Through Equation \ref{calculation}, $M$ can extend the distribution it has learned for $V^{n}$ to one over $V^{n+m}$ for any string of finite length $n+m$. 
Recall that we are looking at strings of ${\cal L}$ that define structures; the structures defined by strings of length $n+m$ are independent of those defined in $V^{n}$ and none is intuitively more likely than another.  So the prior distribution over $V^{n+m}$ should consider as equally likely all continuations $s.a \in V^{n+m}$, where $s \in V^{n}$, $a \in V^m$ and $.$ is concatenation. There are also correspondingly more hypotheses in ${\cal H}^{n+m}$ than in ${\cal H}^n$, since there are $V^{|m|}$ more strings in $V^{n+m}$ than in $V^n$. Thus  $\mu^{n+m}(s.a |h_k) < \mu^n(s.a|h_k)$ for $s.a \in V^{n+m}, s\in V^{n}$ for each $h_k$.  
Non-degenerate distributions will reflect this and should make the model converge to the least general hypothesis supported by the evidence \citep{muggleton:feng:1992,plotkin:1972}.  

\begin{definition} \label{non-degen}
$M$'s distributions over sets of hypotheses ${\cal H}^n$, $\mu^n({\cal H}^n)$, after training over $V^n$ are {\em non-degenerate} if $\forall h \ \forall \delta \ (0 <\delta \leq 1), \exists m>0$ such that $\forall a \in V^m \ \forall s \in V^n : \ \mu^n(s.a | h) = \max\{0, \mu^n (s|h) - \delta\}$, where $s.a \in V^{n+m}$.
\end{definition}

\begin{prop} \label{auto}
Models that calculate distributions over strings using Equation \ref{calculation} have non-degenerate distributions.
\end{prop}
\noindent
As continuations get longer the probability of the continuation will decrease monotonically.$\Box$
Because quantifiers like {\em every} and {\em some} are eliminable in terms of Boolean functions when we consider finite structures definable with strings in $V^*$, we must consider strings in $V^\omega$ to define countably infinite models that capture the full truth conditions of {\em every}.  
To extend a distribution over $V^n$ for finite $n$ to a distribution over $V^\omega$, we lift the probability of a string to the set of its continuations.
   In $V^\omega$, the set of strings $A$ characterizes the set $A.V^\omega$, where  $A.V^\omega$ 
is the set of all strings formed by concatenating a string from $A$  with a string from $V^\omega$.  Using this correspondence, 
the probabilities of sets of finite strings in $V^{n}$ can lifted to probabilities of sets of the form $V^{n}.V^\omega$. The laws of probability extend the distribution to complements, intersections and unions of such sets. 

We now propose a simple but general notion of inductive learning.
\begin{definition} \label{learn} Suppose $\mu_0$ is $M$'s Max Ent prior distribution and let  $h \in {\cal H}^\beta$ for some countable $\beta$.   $M$ {\em effectively learns} $h$ iff after some finite amount of training using inductive inference, there is an $\alpha$, such that: for any $s \in V^{\beta}$, $\mu^\beta(s | h) > \alpha > \mu_0(s |h)$ iff $s \in h$.  
\end{definition}


\begin{prop} \label{compactness}
If $M$ can effectively learn $h^n_\forall$ from sequences of $V_{\cal L}^n$ for arbitrarily large $n \in \omega$, then $M$ can effectively learn $h^\omega_\forall$
\end{prop}
\noindent
  Assume that $M$ cannot effectively learn $h^\omega_\forall$ but it can effectively learn $h^n_\forall$ for arbitrarily large $n \in \omega$. Then it must admit some string $s \in V_{\cal L}^\omega$, such that $s \not\in h^\omega_\forall$. But then at some finite stage $i$, $s_i$ must have $\neg blue(a_i)$.  By hypothesis $M$ has learned $h^i_\forall$.  So it has ruled out $s_i$ and {\em a fortiori} $s$. $\Box$

We now negatively answer our question, \textit{Can $M$ inductively learn hypothesis $h^\omega_\forall$?}, under either of two conditions: (i) $M$ has non-degenerate distributions; (ii) $M$ obeys Max Ent and inductive inference. 

\begin{prop} \label{univ}
Suppose $M$'s distributions are non-degenerate.  Then $h^\omega_\forall$ is not effectively learnable by $M$ over ${\cal H}^\omega$. 
\end{prop}
\noindent

Suppose $M$ trained on $V^{\leq n}$ has effectively learned $h^n_\forall$. So $\forall s \in h^n_\forall$, $\mu^n(s | h_\forall) > \alpha$  where $\alpha$ is as in Definition \ref{learn}. Since $M$'s distributions are non-degenerate, $\exists m$, such that for all $s \in V^m, \exists \delta: 0<\delta \leq 1$ where $\mu^m(s | h_\forall)-\delta < \alpha$ and a  continuation of $s$, $s.a$, such that $s.a \in h^{m+n}_\forall$ but $\mu^{m+n}(s.a |h_\forall) = \mu^m(s |h_\forall)-\delta < \alpha$.  $\Box$

By Propositions \ref{auto} and \ref{univ}, a basic auto-regressive model cannot learn $h_\forall$.  We can generalize Proposition \ref{univ} to other :  
\begin{prop}\label{univ1} Suppose $M$'s priors only obey Max Ent and $M$ uses inductive inference.  Then $h^\omega_\forall$ is not effectively learnable by $M$ over ${\cal H}^\omega$. 
\end{prop}
 Suppose $M$'s training data $D^{\leq n} \subseteq V^{\leq n} $ and $M$ has learned $h^n_\forall$ . To learn $h^{n+m}_\forall$, $M$ must project its distribution of ${\cal H}^n$ onto ${\cal H}^{n+m}$.  But the distributions in ${\cal H}^n$ and ${\cal H}^{n+m}$ are {\em independent}; for one thing the cardinality of ${\cal H}^n$, $|{\cal H}^n|$, is such that $|{\cal H}^n| < |{\cal H}^{n+m}| = |{\cal H}^n|\times 2^m$. 
 Our assumptions about inductive inference on $D^{\leq n}$ make it no more likely that {\em every} will be associated with $h_\forall$ than it is with any of the $2^m$ $h \in 
 {\cal H}^{n+m}$, where strings in $h$ contain the same $n$ prefix as an $s \in h^n_\forall$ but $h \cap h^{n+m}_\forall = \emptyset$.  In ${\cal H}^{n+m}$ these hypotheses $h$ can be distinguished from $h_\forall$. Max Ent priors over ${\cal H}^{n+m}$ imply that for any $s \in V^{n+m}$, $\mu^{n+m}(s | h_\forall) = \frac{1}{2^m} \mu^{n}(s|h_\forall)$. 
 $\Box$

 \begin{cor} \label{finite-basic}
$M$ cannot effectively learn $h^n_\forall$ from sequences in $V_{\cal L}^n$ for arbitrarily large $n \in \mathbb{N}$.  There is some $n$ such that $h^n_\forall$ is not effectively learnable.
\end{cor}

While LLMs can represent any Borel function to an arbitrary degree of precision \cite{hornik:etal:1989}, Propositions \ref{univ} and \ref{univ1}  shows they cannot always learn such functions, given either the constraints of inductive epistemology or the way LLMs generate probabilities for strings. In particular, given our assumptions, no LLM can effectively learn $h^\omega_\forall$.  In addition, each LLM is bounded by some number $n$ in the size of sequences for which it can learn $h^n_{\forall}$.  LLMs do not have the capacity to learn the meaning of `every' even over all finite domains.\footnote{Unlike  Hume's problem of induction \citep{popper:1963} and \citep{wolpert:macready:1995}, we exploit particularities of LLMs and the structure of a classification problem. The finite bound on learning of hypotheses goes beyond  standard Humean conclusions}  

Even supposing that an LLM can effectively learn $h^n_\forall$ for some $n$, this does not amount to understanding {\em every}.  $h^n_\forall$ can be effectively represented with quantifier free conjunctions of formulae, and these do {\em not} correctly approximate reasoning with a sentence like \ref{blue} that applies to arbitrarily large domains. Identifying $\forall$ with a finite conjunction of length $n$ will make $\forall x Fx$ consistent with $\neg \forall x Fx$ in larger structures.  In $\omega$ structures, for example, $\neg \forall x Fx$ is consistent with every finite subset of the $\Pi^0_1$  string $\mathit{blue}(0), \ \mathit{blue}(1), \ \mathit{blue}(2), \ \ldots$, in $h^\omega_\forall$, making it inevitable that LLMs will reason incorrectly with {\em every} in large enough structures.  

The situation worsens with sampling: suppose that when we present our model $M$ a long string, $M$ only samples some of the elements in the string; the threat of inconsistency in such a situation can become high and we have no guarantees that such inconsistencies will not arise.\footnote{Approximation and approximation error can also affect learnability of mathematical functions \cite{colbrook:etal:2022}.}  But this reasoning is not {\em independent} of the meaning of {\em every}; as the semantics and rules of first order logic show, this reasoning is {\em an integral part} of the meaning of {\em every}.  As a result, LLMs unable to grasp semantic consequence defined in terms of universal quantification; and we thus cannot provide them guarantees that they follow semantic entailments when asked to do semantic tasks.  This predicts phenomena like LLM hallucinations and observed elementary reasoning errors.

\subsection{Generalizing our answer to Q2}\label{subsec:general}

Using tools from statistical learning and the Borel Hierarchy, we now generalize Propositions \ref{univ} and \ref{univ1} to other concepts beyond {\em every}. 

\paragraph{Statistical learning} examines the application of a learned function over a test domain and the expected loss over novel applications.  The ability to bring the error over test to that over the training set is typically taken to indicate an ability to generalize \cite{neyshabur:etal:2017general}.  
\noindent
\citet{villa:etal:2013} define learnability in statistical learning theory via the notion of {\em uniform consistency}.  Let $\mu$ be a distribution over ${\cal H}$ and $\mu_n$ the update of $\mu$ after $n$ training samples $z_i = (x_i, y_i)$. Let $A_{z_n}$ be an algorithm for picking out a hypothesis from ${\cal H}$ based on $n$ training samples.  $\mathit{inf}_{\cal H}$ is the hypothesis in ${\cal  H}$ with the lowest possible error \cite{shalev:etal:2010,kawaguchi:etal:2017generalization}.
\begin{definition} \label{uniform1}
An algorithm $A$ on a hypothesis space ${\cal H}$ is uniformly consistent if and only if\\
$\forall \epsilon > 0  \  lim_{n \rightarrow \infty}
\mathit{sup}_\mu $\\
\hspace*{.3in} $\mu_n (\{z_n: \mathbb{E}_\mu(\{A_{z_n} - \mathit{inf}_{\cal H} \mathbb{E}_\mu > \epsilon\}) = 0$
\end{definition}
In our case, the best hypothesis, $\mathit{inf}_{\cal H}$, for instance $h_\forall$, will yield $0$ error.  Our question is whether there is an algorithm that converges to that hypothesis given a certain ${\cal H}$ and certain assumptions.
\begin{definition} A class of hypotheses ${\cal H}$ is {\em uniformally learnable} just in case there exists a uniformly consistent algorithm for ${\cal H}$.
\end{definition}
This enables us to link learnability with a number of other features: 
\begin{theorem} \label{anthony} \cite{anthony:etal:1999} Let $Y = \{0, 1\}$. Then the following
conditions are equivalent: (i)
${\cal H}$ is uniformly learnable;
(ii) Empirical risk minimization on ${\cal H}$ is uniformly consistent;
(iii) ${\cal H}$ is a uGC-class;
(iv) the VC-dimension of ${\cal H}$ is finite.
\end{theorem}

\paragraph{The Borel Hierarchy}
We now turn to generalize the hypotheses we are investigating.  $V^\omega$  has a natural topology, the Cantor topology, which allows us to characterize linguistic concepts precisely.  To define the topology, we first define the basic open sets  to be sets of the form $A.V^\omega$, denoted as $\open(A)$, where $A \subseteq V^*$ is a set of finite strings.  Importantly, $\open(A)$ sets are both open and closed or {\em clopen}, because if $A\subset V^*$ is a countable set, then the complement of $A.V^\omega$, $(V^*\setminus A).V^\omega$, is also open.  And thus, $A.V^\omega$ is also closed. The $\Delta^0_1$ class is at the intersection of the $\Sigma^0_1$ and $\Pi^0_1$ classes and consists of the clopen sets. $\Sigma^0_1$ sets include countable unions of $\Delta^0_1$ sets, while $\Pi^0_1$ are complements of $\Sigma^0_1$ sets and so include countable intersections of $\Delta^0_1$ sets.  

 These sets form the basis of the {\em Borel hierarchy} of sets that includes the $\Delta^0_1$, $\Sigma^{0}_1$, and  $\Pi^{0}_1$ sets, and more generally includes $\Sigma_{\alpha+1}^0$ as the countable union of all
$\Pi_{\alpha}^0$ and $\Delta_{\alpha}^0$ sets, and $\Pi_{\alpha+1}^0$ as the complement
of  $\Sigma_{\alpha+1}^0$ sets, with $\Delta_{\alpha}^0 = \Sigma_{\alpha}^0 \cap \Pi_{\alpha}^0$. The hierarchy is strict and does not collapse~\cite{kechris:1995}.  We will use this hierarchy to characterize linguistic concepts.  Below is a picture of some simple Borel sets and their $\subseteq$ relations.
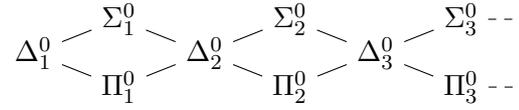
\begin{center}
\begin{tikzpicture}[scale=0.45]
 
  \node (s1) at (0,0) {$\Sigma_1^0$};
  \node (s2) at (5,0) {$\Sigma_2^0$};
  \node (s3) at (10,0) {$\Sigma_3^0$};
  \node (s4) at (12,0) {};

  \node (p1) at (0,-2) {$\Pi_1^0$};
  \node (p2) at (5,-2) {$\Pi_2^0$};
  \node (p3) at (10,-2) {$\Pi_3^0$};
  \node (p4) at (12,-2) {};

  \node (d1) at (-2.5,-1) {$\Delta_1^0$};
  \node (d2) at (2.5,-1) {$\Delta_2^0$};
  \node (d3) at (7.5,-1) {$\Delta_3^0$};

\draw (d1) --(s1);
\draw (d1)--(p1);
\draw (s1)--(d2);
\draw (p1)--(d2);
\draw (d2) --(s2);
\draw (d2)--(p2);
\draw (s2)--(d3);
\draw (p2)--(d3);
\draw (d3) --(s3);
\draw (d3)--(p3);
\draw[dashed] (s3) -- (s4);
\draw[dashed] (p3) -- (p4);
\end{tikzpicture}
\end{center} 
As an example, $h^\omega_\forall \subseteq V_{\cal L}^\omega$ of the previous section is a $\Pi^0_1$ Borel set; i.e., $h^\omega_\forall = \bigcap_{i\in \omega} B_i$ where the $B_i$ are $\Delta^0_1$.

We are interested in the learnability of Borel sets $B$ with respect to a hypothesis space.  The hypothesis space ${\cal H}^n$ for $V^{\leq n}$ and algorithem ${\cal A}^n$ that an LLM can consider is typically fixed by the maximal length strings it has been trained on.   But we will be looking at how an LLM extends its training generalizing to longer and longer strings.  More generally, we consider  a countable collection of hypotheses---in the case of {\em every} and $V_{\cal L}$, the set consists of $h_\forall, h_{{\mathit the first} \ 2^n}$ etc.  We will assume a countable hypothesis space ${\cal H}^\omega$ for the Borel sets in $V^\omega$, with $|V| > 2$ we want to learn in what follows.

\begin{definition}
An LLM  $M$ can effectively learn a Borel set $S \subset V^\omega$ out of a countable set of hypotheses ${\cal H}$ iff $M$ has a uniformly consistent algorithm such that  $h_S = \mathit{inf}{\cal H}$, as in Definition \ref{uniform1}, and where $h_S: V^\omega \rightarrow \{0,1\}$ defines $S$.
\end{definition} 
 Clearly if $h$ is $\mathit{inf}{\cal H}$, and ${\cal A}$ is uniformly consistent, then Definition \ref{learn} is satisfied; i.e., there is some $\alpha > \epsilon$ such that $\mu(s|h) > \alpha$ iff $s \in h$.

 

\begin{theorem} \label{main} An LLM with either (a) non degenerate distribution or (b) Max Ent priors and trained on  $V^{<n}$ for some finite $n$ to learn $h \subset V^\omega$ via inductive inference (i) can effectively learn a $\Delta^0_1$ set $\open(S) \subset V^\omega_{{\cal L}}$, where $S$ is a finite subset of $V^{\leq n}$, given ${\cal H}_{\open(V^{<n})}$, a hypothesis space restricted to $\Delta^0_1$ sets; but (ii) it cannot effectively learn any $\Pi^0_1$ Borel set $B \subset V^\omega_{{\cal L}}$.
\end{theorem}
\noindent
We first show (i).  Let ${\cal H} = \{\open(A) : A \subseteq V^{<n}\}$.  Any $h \in {\cal H}$ is determined by a finite set of prefixes $P$ in $V^{<n}$.  There are only finitely many such sets in $V^{\leq n}$, and so $M$ has an algorithm $A$ that eliminates at each finite stage of training some $\Delta^0_1$ $\open(P)$ sets.  This enables it to converge uniformly toward $0$ expected error for the set of finite prefixes that determines $\open(S)$ and so eventually $M$ will have effectively learned $\open(S)$.

Now for (ii). We first consider the case  (ii.a) where our learned model has non-degenerate distributions.  Consider an arbitrary $\Pi^0_1$ complete set $B$. So $B  = \bigcap_{n \in \omega} \open(B_n)$, with $\open(B_{n+1}) \subset \open(B_n)$, where the $B_i \subset V^*$. To compute $B$, $M$ needs a uniformly consistent algorithm ${\cal A}$ over our countable hypothesis space ${\cal H}$ that converges on $h_B$, the hypothesis defining $B$. Now suppose $M$ has been trained on strings in $V^{<n}$; its algorithms ${\cal A}$ are thus restricted to ${\cal H}^{<n}$.  

Suppose $M$ trained on $V^{\leq n}$ has effectively learned $h^n_{B_n}$. 
Let $s \in h^n_{B_n}$.  Since $M$'s distributions are non-degenerate, $\forall \alpha \geq 0, \exists m$, $\delta: 0<\delta \leq 1$ where $\mu^m(s |h_B)-\delta < \alpha$ and a  continuation of $s$, $s.a$, such that $s.a \in h^{m+n}_{B_n}$ but $\mu^m_M(s.a |h_B) = \mu^m_M(s |h_B)-\delta < \alpha$.  
So there is no convergence at any finite stage $n$ of ${\cal A}$ towards $h_B$.  Non uniform learnability of ${\cal H}$ then follows. 

(ii.b) Let's now assume that $M$ only has Max Ent priors and learns by inductive inference. Uniform convergence of any algorithm obeying these conditions is not guaranteed as a similar argument as in Proposition \ref{univ} applies.  $\Box$ 

\begin{cor} \label{uniform}
The hypothesis space ${\cal H}_B$ is not uniformly learnable.   Hence the the VC-dimensions of  ${\cal H}_B$ are not finite, and empirical risk minimization on ${\cal H}_B$ are not uniformly consistent.
\end{cor}
\begin{cor} \label{sigma}
$M$ cannot effectively learn $\Sigma^0_1$ complete Borel sets.
\end{cor}
\noindent
Assume $M$ can effectively learn a $\Sigma^0_1$ complete set.  Then it can effectively learn a $\Pi^0_1$ set that is its complement, which is impossible by Theorem
\ref{main}.$\Box$ 

\begin{prop} \label{higher}
An LLM $M$ cannot effectively learn Borel sets $B$ of higher complexity than $\Delta^0_1$.
\end{prop}
\noindent
Proposition \ref{main} and Corollary \ref{uniform} show that \ $M$ cannot effectively learn $\Pi^0_1$ or $\Sigma^0_1$ sets.  But any $\Pi^0_n$ or $\Sigma^0_n$ complete Borel set $B$ for n > 1 is at least a countable intersection or countable union of such sets.  So $B$ is not effectively learnable.$\Box$

\citet{JPL,JOLLI} 
%
 examine concepts of discourse consistency and textual and conversational coherence, which true, human-like conversational capacity requires. Using continuations in a game-theoretic setting, they show those concepts determine more complex $\Pi^0_2$ sets in the Borel Hierarchy; and intuitive measures of conversational success---like the fact that one player has more successful unrefuted attacks on an opponent's position than vice versa---determine $\Pi^0_3$ sets.  Given Proposition \ref{higher}, LLMs cannot learn these concepts, which are needed for full conversational mastery.


\begin{prop} \label{finite}  For any LLM $M$, there is a maximally large and fixed number $n$ such that ${\cal H}^n$ is uniformly learnable for $M$ but ${\cal H}^{n+k}$ is not uniformly learnable, for $k > 0$. 
\end{prop}
\noindent
Suppose that for $M$ ${\cal H}^n$ is uniformly learnable for all $n$.  Then, $M$ can compute the countable intersection of sets defined by the best hypotheses in ${\cal H}^n$ for each $n$.  So $M$ can effectively learn a $\Pi^0_1$ set, which contradicts Theorem \ref{main}.$\Box$ 
\begin{cor} \label{finite1}
 $M$ cannot effectively learn $\Delta_1^0$ sets of the form $\open(A)$ if the length of $A$ is longer than the maximal number $n$ such that ${\cal H}^n$ is uniformly learnable for $M$.
\end{cor}
\subsection{The importance of order}
Order is important for the most elementary reasoning about linguistic content in finite domains.  
Let us add another predicate $A$ to ${\cal L}$ to form the language ${\cal L}^+$. Now consider the strings in $V^\omega_{{\cal L}^+}$.  Strings consistent with \ref{blue} may include formulae like $A(a_i)$ or  $\neg A(a_i)$, paired with a choice of $\mathit{blue}(a_i)$ or $\neg \mathit{blue}(a_i)$. Even to find effectively initial segments of strings in $h^\omega_{\forall_{{\cal L}^+}}$, $M$ must learn some sentence structure or word order.  The negation sign has to be paired with the predicate {\em blue}; if it's appended to $A$ (e.g., \textit{large}, or some other independent term), this should count as a string in $h^n_\forall$.  If $s$ is a finite string,  $M$ does not effectively capture word order if it does not distinguish between $s$ and permutations of elements in $s$.
\begin{prop} \label{prop}
If $M$ does not effectively capture word order, it cannot effectively learn basic sets of the form $\open(A)$ for $A \subset V^*$.
\end{prop}
\noindent
Let $s \in A$ be a string containing $A(a_i) \wedge \neg \mathit{blue}(a_i)$ but $A$ has no string containing $\neg A(a_i) \wedge \mathit{blue}(a_i)$.  If $M$ does not capture word order, $M$ cannot distinguish between $s$ and $s$'s permutation containing $\neg A(a_i) \wedge \mathit{blue}(a_i)$.  
$\Box$

\begin{cor}
 If $M$ does not effectively capture word order, it will not reason soundly in propositional logic. 
\end{cor}
The example in Proposition \ref{prop} shows that $M$ will not be able to reason about logical structure if it does not effectively capture word order. $\Box$

\citet{yuksekgonul:etal:2022,sinha:etal:2020unnatural} provide evidence that small to moderate sized LLMs do not reliably capture word order. Our empirical examples show even GPT3.5 and ChatGPT have difficulties with sentential word order, and, worryingly, with the order of arguments in a logical operator; the example in Appendix B suggests that even ChatGPT can't be trusted to always do elementary inferences involving conditionals correctly. 
Thus, LLMs with their initial training do not necessarily find basic $\Delta^0_1$ sets of the form $a.V^\omega$ where $a$ is a single string but only sets $A.V^\omega$ where $A$ is a set of prefixes that are permutations on $a$.  This is surprising and poses extreme difficulties for valid reasoning with operators that have order dependent arguments.  

\section{Empirical investigations of LLMs with {\em every}}\label{sec:q1}
While the theoretical argument laid out in Section \ref{sec:limits} does not hinge on empirical statistics of LLM failures, it certainly suggests that we should expect such failures. In this section, we describe some of the tests we have performed using continuations to query LLMs directly about their mastery of universal quantification.


Let us return to our simple example from above, repeated here as \ref{blue}:
\ex. Everything is blue. \label{blue}   

We used finite sequences of formulas as a context, e.g.,  $a_{1} \ \mathit{ is \ blue}$, $a_{2} \ \mathit{is \  red}$, $a_3 \mathit{\ is\ red}$,..., $a_{i} \ \mathit{is \ blue}$ to determine a model.  We then asked an LLM $M$ whether \ref{blue} in this model,
allowing us to gauge its behavior with respect to finite domains. 

BERT-large and RoBERTa-large already failed to reliably distinguish very small models (containing 2 and 5 objects respectively) in which \ref{blue} is true from those in which it is not.
   To test these models, 
we fine-tuned BERT-large and RoBERTa-large on the CoQA dataset \cite{coqa}. For finetuning, the model had $4$ output heads for \emph{yes}, \emph{no}, \emph{unknown}, and \emph{span} type questions. Since the CoQA dataset provides a rationale for each question, the models were jointly trained on question answering and rationale tagging tasks to enhance their performance. We report scores on the finetuned models on CoQA for $1$ epoch as we did not observe significant improvement with an increased number of epochs.

For BERT-large, we provided strings like \ref{simple1} and then asked \textit{Is everything blue?}
\ex.
\label{simple1} My car is blue. My house is blue

We generated a total of $5$ examples in which $\ref{blue}$ was true  and $5$  examples in which \ref{blue} is false. All the examples had only $2$ objects.  The inconsistent examples were constructed by varying the position of the object which was inconsistent with the asked question and by trying out different combination of colours and objects. 

\vspace{3mm}
\noindent The consistent examples were of the form: \\
1. The car is blue. The house is blue. \\
2. The car is purple. The house is purple. \\
3. The car is yellow. The house is yellow. \\
3. The shirt is violet. The table is violet. \\
4. The cup is black. The plate is black.
    
\vspace{3mm}    
\noindent Inconsistent examples were of the form: \\
1. The car is blue. The house is red. \\
2. The car is green. The house is purple. \\
3. The car is yellow. The house is brown. \\
2. The shirt is violet. The table is brown. \\
3. The cup is black. The plate is white. 

\vspace{3mm}
BERT-large was able to correctly identify the consistent examples but failed for all the inconsistent examples.  As the model failed for all the inconsistent examples with $2$ objects, we did not experiment with models containing more than $2$ objects.

For RoBERTa-large, we generated a total of $9$ consistent examples and $53$ inconsistent examples. We constructed sequences ranging from $2$ to $10$ objects. For each number, the inconsistent examples were constructed by varying the position of the object in the string (context) which is responsible for the inconsistency. The model was able to correctly identify all the consistent examples.  For models of a given size (i.e., number of objects), we defined the {\em  pass fraction} as the ratio of the examples in which the model was able to report models inconsistent with \ref{blue} correctly to the total number of inconsistent examples. Table~\ref{tab:pf-roberta} reports the pass fraction on inconsistent examples. 
\begin{table}
\centering
\footnotesize
\begin{tabular}{||c c||}
 \hline
 Object Count & Pass Fraction \\ [0.5ex] 
 \hline\hline
 2  & 1/1 \\ 
 \hline
 3 & 2/3 \\
 \hline
 4 & 1/4 \\
 \hline
 5 & 0/5 \\
 \hline
 6 & 0/6 \\  
 \hline
 7 & 0/7 \\
 \hline
 8 & 0/8 \\
 \hline
 9 & 0/9 \\
 \hline
 10 & 0/10 \\[1ex]
 \hline
\end{tabular}
\caption{Pass fraction on inconsistent examples for RoBERTa-large}
\label{tab:pf-roberta}
\end{table}

While BERT's and RoBERTa's behavior was stable on the strings tested, GPT3.5 davinci and ChatGPT, while more robust, are unstable from one day to the next, even when temperature is set to 0 (on GPT3.5). This made it difficult to pin down the models' abilities, though some generalizations emerged. Typically (though not always), these models can recognize which objects in a string have a certain property, but they cannot necessarily exploit this information to answer 
questions about the string as a whole (see the ``hats'' example in Appendix A).    
In addition both GPT3.5 and ChatGPT will sometimes (frequently in our most recent tests) over-generalize and say that all items in a list are, say, blue if it is specified for all items but one that they are blue and it is not specified one way or the other for the remaining item (see the \textit{fifteen hearts} example from ChatGPT in Appendix A). Thus, even these sophisticated models still fail on more complicated questions and longer strings. 


Our empirical observations on LLMs like BERT and RoBERTa and probing of ChatGPT strongly support our argument that LLMs are unable to master quantification, complementing   
observed LLM difficulties with negation \cite{naik-etal-2018stress, kassner:schutze:2019, hossain-etal-2020-analysis,hosseini:etal:2021} and to some extent quantifiers \cite{kalouli:etal:2022}.  

\section{Conclusions}\label{sec:conclusion}
 We have shown that LLMs' demonstrably inadequate grasp of the meanings of words like {\em every} and other linguistic constructions has a theoretical foundation and explanation:  for certain expressions $S$, $S$'s content should be defined via consistent sets of strings in $V^\omega$, and  LLMs cannot effectively learn certain sets in $V^\omega$. More generally, LLMs cannot effectively learn full meanings of first order quantifiers or any Borel sets beyond the basic open sets, which means that they fail to grasp the meaning of a long list of mundane, frequently used expressions.
 
  Many of these expressions are syncategorematic terms and express what we might call {\em precise concepts}.  Such concepts are needed for understanding ordinary entailment across all expressions; in addition, correctly reasoning with these concepts and grasping their entailments is essential to understanding them.  Reasoning and entailment are intimately tied with meanings.  For us and most formal semanticists \cite{montague:1974}, grasping  meaning and correctly reasoning with linguistically expressed concepts go hand in hand; if you cannot exploit the meanings of words in correct reasoning, you do not really know  what they mean.  The incorrect reasoning of LLMs exemplifies their failure to grasp semantic entailments and meaning.

Our arguments go beyond those of \citet{bender-koller-2020-climbing}, who argue that stochastic models cannot capture linguistic meaning because they consider only form, not denotation. While we agree that denotation plays a very important role in meaning for many expressions, the meaning of most expressions, and especially that of syncategoregmatic ones, requires us to capture their semantic entailments. We have shown that we can capture these entailments within the semantic framework of LLMs using continuation semantics.  But we have also shown that LLMs nevertheless fail in this task. 


LLMs \textit{can} learn certain types of $\Delta^0_1$ sets and finite intersections and unions of learnable $\Delta^0_1$ sets.  For many open class words---including many nouns, adjectives and verbs---whose characteristic denotations can be determined given a finite sample, this probably suffices to capture their meaning or at least a very good approximation of it.  In addition, many NLP tasks may not involve logical inference but an independent form of string optimization; in text summarization or translation, where given a context $s$, $M$ tries to find an optimal continuation $s'$.  If the length of $s.s'$ falls within the constraints of Corollary \ref{finite1}, then we can expect an LLM to succeed at such a task.

Proposition \ref{finite} and Corollary \ref{finite1} generalize Corollary \ref{finite-basic} and they all point to a general limit on learnability for LLMs.   They establish that language models have strict bounds even on the $\Delta^0_1$ sets they can effectively learn.  So we cannot count on LLMs having full linguistic competence even on finite domains.  Different models may have different limits; smaller models generally with lower limits.  This motivates a comparative study of the limits of learnability for different LLMs, complementing  \citet{colbrook:etal:2022}.


Because we do not make  assumptions about memory but only about inductive processes and learning,  our results hold for arbitrarily large LLMs and for any task that relies on an LLM's capacity of string prediction, even if strings are not directly predicted.  

Our research implies that full language mastery needs a different approach from one in which one seeks to build ever larger LLMs with language masking or autoregressive training.  Following \citet{raissi:etal:2017pinns}, we believe we need to inject knowledge about linguistic structure and content into our models  to further constrain learning and in particular hypothesis sets and learning architectures as suggested in \citet{steinert:etal:2019}.

\newpage
\section*{Limitations}

Our paper offers an in-depth look at the limits of statistical learnability for LLMs in terms of Borel sets.  Our results however rely on certain, rather mild hypotheses about learning. We do not know how our claims fare when different assumptions about learning are made.   

Another limitation is that we have not looked at potential ways of restricting learning hypotheses about the meanings of linguistic expressions.  Directly relevant are the constraints on generalized quantifiers of \citet{barwise:cooper:1981}.
Nor have we said much about compositionality and learning, although they are well known to interact.   Ideally, we should assign a meaning to {\em every} which consists in a reduced hypothesis space, but that has to combine with the meanings of, say, the other words in \ref{blue}.  

A third limitation concerns the extent to which an LLM can leverage what it has been trained to do in learning semantic concepts on other tasks.  In many experiments we see that an LLM can appear to learn X but then cannot exploit X in task Y.   We leave this issue for future research, but we believe it may be key to understanding debates about LLM performance and the role of forgetting in LLM learning \cite{kirkpatrick:etal:2017}.

\section*{Ethics Statement}
 We show that LLM reliability in gauging the accuracy of universal quantifiers is limited.  Given how quantification is linked to concepts like semantic consequence, it means that we cannot assume LLMs to have anything like full semantic competence.  This suggests that we cannot fully trust them with tasks where accuracy on reasoning is required.  Thus, our work has certain ethical implications and provides a cautionary tale for those working on LLMs as general AI models.

 \section*{Acknowledgments}
For financial support, we thank the National Interdisciplinary Artificial Intelligence Institute 
ANITI (Artificial and Natural
Intelligence Toulouse Institute), funded by the French ‘Investing for 
the Future– PIA3’ program under the Grant agreement ANR-19-PI3A-000, and we also thank the projects COCOBOTS (ANR-21-FAI2-0005) and DISCUTER (ANR-21-ASIA-0005).  We also thank the COCOPIL ``Graine" project of the Région Occitanie of France.

\bibliography{multimodal}

\appendix

\


\section{GPT 3.5 and ChatGPT on {\em every}}

Because our access to GPT 3.5 and ChatGPT is limited and the responses that we receive seem to vary depending on the day that we query the models, we qualify the results in these appendices as anecdotal. Nevertheless, we emphasize that for each type of example that we provide, we made numerous trials in order to make sure that the results were repeated and so seemed to follow from features of the model, not just random quirks.

In the queries below, GPT3.5 on 18.01.2023, with temperature set to 0, correctly classifies both blue hats as being blue when queried directly about the color of the beret, but still fails to correctly answer \textit{Were all of the hats blue?} (Results sometimes varied with the form of the prompt; e.g., using commas rather than \textit{and} can impact GPT's answers).  

\noindent
Input Text: One guy was wearing a blue beret and another guy was wearing a pink bonnet and a third guy was wearing a blue baseball cap. 

\vspace{1.5mm}
\noindent
Q: Were all of the hats blue?\\
GPT: No, only one of the hats was blue.

\noindent
Q: Which hat was blue?  GPT: The blue baseball cap.

\noindent
Q: Is a beret a hat?  GPT: Yes, a beret is a type of hat.

\noindent
Q: What color is the beret?  GPT: The beret is blue.

\vspace{.2in}
Here is an input and query to ChatGPT on 20.01.2023, showing that it ignores underspecification

\noindent
Input text: I have a shirt with a picture of ten blue bears, five blue snails, three blue guitars, fifteen hearts, 15 blue fluffy rabbits and 25 blue stars. Is everything on my shirt blue?

\vspace{1mm}
\noindent
ChatGPT: Yes, everything on your shirt is blue.

\section{ChatGPT and GPT3.5 on word order and argument order}

The examples below illustrate how both GPT3.5 and ChatGPT fail to recognize instructions about word order. When asked to write sentences that end with a particular word, for example, the models manage to produce sentences in which the target word appears, but in most cases, it is not in the correct position. 

Instruction: Write five sentences that end with the word ``house''. \\

\noindent {\bf ChatGPT}: \\
1. I built a house with my own two hands. \\
2. The storm destroyed the old house. \\
3. We found a beautiful house to rent for the summer. \\
4. The house was haunted, so we left in the middle of the night. \\
5. I always dreamed of living in a big white house. \\

\noindent {\bf GPT 3.5}: \\
1. I can't wait to get home and relax in my own house. \\
2. The old house had been abandoned for years. \\
3.  The house was filled with the smell of freshly baked cookies. \\
4. The house was surrounded by a tall white picket fence. \\
5.  The house was the perfect place to spend a lazy summer day.\\

For GPT3.5  we set temperature to $0$. These observations suggest both GPT 3.5 and ChatGPT fail to respect the {\em last word} constraint.

Finally, we offer this example that shows that ChatGPT can confuse the order of arguments in a logical operator, thus leading to incorrect reasoning.

\vspace{3mm}
\noindent
Prompt: If p then q. q, therefore p. \\

\noindent {\bf ChatGPT}: \\
This is a valid logical argument. It is known as modus ponens and it is a valid rule of inference in propositional logic. The statement ``If p then q'' is known as the conditional statement and ``q, therefore p'' is the conclusion that can be drawn from the conditional statement.

\label{gpt-logic}

\end{document}